# Dynamic Modeling and Simulation of a Four-wheel Skid-Steer Mobile Robot using Linear Graphs

Eric McCormick, Haoxiang Lang, *Senior Member, IEEE,* and Clarence W. de Silva, *Fellow, IEEE*

*Abstract*— This paper presents the application of the concepts and approaches of linear graph (LG) theory in the modeling and simulation of a 4-wheel skid-steer mobile robotic system. An LG representation of the system is proposed and the accompanying state-space model of the dynamics of a mobile robot system is evaluated using the associated LGtheory MATLAB toolbox, which was developed in our lab. A genetic algorithm (GA)-based parameter estimation method is employed to determine the system parameters, which leads to a very accurate simulation of the model. The developed model is then evaluated and validated by comparing the simulated LG model trajectory with the trajectory of a ROS Gazebo simulated robot and experimental data obtained from the physical robotic system. The obtained results demonstrate that the proposed LG model, combined with the GA parameter estimation process, produces a highly accurate method of modeling and simulating a mobile robotic system.

## I. Introduction

Engineering systems are becoming increasingly more advanced with the integration of multiple physical domains such as mechanical, electrical, fluid and thermal into system in an integrated and unified manner, which may have traditionally only been considered as an interconnected independent subsystems, each of which comprised just a single physical/energy domain. The resulted field of multi-disciplinary engineering, known as mechatronic engineering, has brought with it a significant acceleration in technological advancement. While this advancement has come with parallel advancements in computer-automated robotic, vehicular, machine tool, and electronic systems, and more sophisticated control, this has also resulted in an increase in the complexity associated with modeling, simulating, designing, and controlling such multi-domain engineering systems. In recent decades, many methods of system modeling have been introduced to address this complexity issue, but most of them lack the necessary integrated and unified focus of Mechatronics, which can be facilitated by graph-based modeling methods. In particular, Henry M. Paynter in the 1960s, introduced two separate graph-based methods of dynamic system modeling, linear graphs (LGs) and bond graphs (BGs) [1].

While BG modeling has seemingly surpassed the popularity of LG modeling, due in part to the development of commercial software tools such as 20sim, which facilitate the process of evaluating complex BG models [2, 3, 4, 5], the LG approach provides some additional benefits beyond those associated with the BG theory.

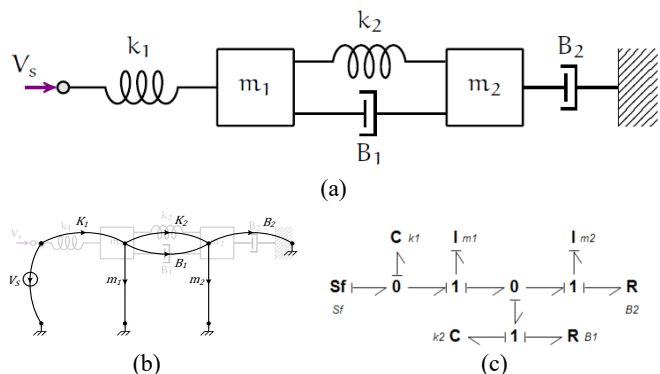

Fig. 1. (a) Schematic model of a Mass-Spring-Damper System; (b) Equivalent LG model; (c) Equivalent BG model.

The key benefit that the LG approach provides over the BG approach is the intuitive nature in which LG models can be constructed and recognized so that the model structure directly corresponds to the actual physical structure of the system. Fig. 1 shows an example of modeling a mechanical system using LG and BG, respectively. Notably, electrical and mechanical systems will have analogous structures in their LG models. For many energy domains, such as electrical and fluid, the conversion of their schematic diagrams to LG models can often result in a nearly identical topological layout.

This work was supported by Natural Science and Engineering Research Council of Canada (RGPIN-2017-05762).

Eric McCormick is with the GRASP Lab at Ontario Tech University, Oshawa, ON, Canada (e-mail: eric.mccormick@ontariotechu.net).

Haoxiang Lang is with the GRASP Lab at Ontario Tech University, Oshawa, ON, Canada (e-mail: haoxiang.lang@uoit.ca).

Clarence W. de Silva is with the Industrial Automation Laboratory at The University of British Columbia, Vancouver, BC, Canada (e-mail: desilva@mech.ubc.ca).

Additionally, the use of simple node and loop network equations, like those of Kirchhoff's current and voltage laws, allow for the analogous application of electrical network-like algorithms across systems consisting of many energy domains, leading to analogous loop equations and analogous node equations. Similarly, the network representation provided by the LG approach provides a more intuitive method of determining the power flow of the system, thus allowing for an easier method of determining the dependent and independent variables of the system. Furthermore, LG modules can be introduced to represent such physical devices as amplifiers, by means of modulated source elements, converters from the energy domain into information domain (e.g., process to sensor), and converters from the information domain into the energy domain (e.g., control to actuator) [6].

LG theory, evolved from Leonhard Euler's graph theory, was first applied in engineering for the analysis of large electrical networks before being expanded to applications spanning multiple energy domains [7]. Before LG theory, analysis of multi-domain systems was neither integrated nor unified, meaning that, different physical domains were modeled separately without considering their dynamic interactions, and different (not analogous) techniques were used for modeling each domain. While initial developments in LG theory focused on the primary energy domains (electrical, mechanical, fluid, and thermal), developments in additional domains, such as thermohydraulic [8, 9], electrochemical [10], and multibody [11, 12], started to emerge. While the recent work in the field of LGs is rather limited when compared to BGs, some research is being conducted into the application of LG theory for automatic design evolution [13].

To demonstrate the LG approach and its versatility and robustness, this paper develops an LG model representation of the dynamics of a four-wheel skid-steer mobile robot and verify the accuracy by comparing the physical system and existing model provided in the popular robotics simulator (Gazebo). A MATLAB toolbox named the LGtheory MATLAB toolbox has been developed in our laboratory, to address the lack of available LG-based software tools [14]. This toolbox will be utilized to automate the process of formulating the state-space model of the dynamic system. This model will be further enhanced with a genetic algorithm (GA)-based parameter estimation procedure in order to calibrate the unknown parameter values of the model. Evaluation and validation of the proposed model will be performed via comparisons of the trajectory response of the simulated LG model against the data collected from a Robot Operating System (ROS)-based Gazebo simulation and from experimental data of real-world driving scenarios using a Clearpath Husky mobile robotic system.

## II. LINEAR GRAPH MODELING

The LG approach is a method of modeling and evaluating complex dynamic systems through the use of a simplistic graphical representation in order to derive their state-space models. This is a systematic, unique, integrated, and unified approach, which uses well-established set of steps (systematic) leading to a single model (unique), by considering all physical domains and their interactions simultaneously (integrated), and using analogous methods to model the different domains (unified). Essentially, it provides a robust modeling method, which produces a single unique model for a specific system through the analogous application of methodologies across multiple energy domains. This means that a multi-domain mechatronic system is evaluated using a single integrated model, not as a series of separate models (which is the "sequential" approach), while applying similar network equations and algorithms for each of the system's energy domains [6].

The procedure introduced in [15] can be applied into the LG representation of the physical model to derive a standard form of state-space model of the modeled multi-domain engineering system as shown in Equation (1):

$$\dot{x} = Ax + Bu$$
$$y = Cx + Du \quad (1)$$

The conversion between the graphical LG model and the mathematical state-space model is achieved by using the developed LGtheory Matlab Toolbox in our lab [14].

Fig. 2 shows an example of modeling a hydro-mechanical system consisting an electric motor to power a positive-displacement pump and piston which actuate a mass element attached to a spring and to ground in the mechanical translation domain.

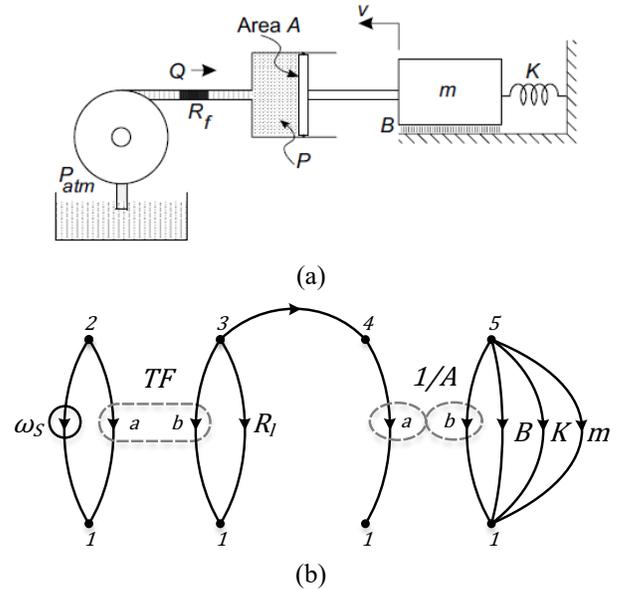

Fig. 2. (a) System model and (b) LG model.

The following equations represent the state-space model produced by the toolbox with outputs specified as the pressure of the fluid between the pump and the piston, and the velocity of the mass element:

$$\begin{bmatrix} \dot{v}_m \\ \dot{F}_k \end{bmatrix} = \begin{bmatrix} \frac{A^2(R_f + R_l - R_f R_l TF)}{m(R_l TF - 1)} - \frac{B}{m} & -\frac{1}{m} \\ K & 0 \end{bmatrix} \begin{bmatrix} v_m \\ F_k \end{bmatrix} + \begin{bmatrix} 0 \\ 0 \end{bmatrix} [\omega_s]$$
$$\begin{bmatrix} P_{R_f} \\ v_m \end{bmatrix} = \begin{bmatrix} -AR_f & 0 \\ 1 & 0 \end{bmatrix} \begin{bmatrix} v_m \\ F_k \end{bmatrix} + \begin{bmatrix} 0 \\ 0 \end{bmatrix} [\omega_s] \quad (2)$$

With the successful development and validation of an automated tool for evaluating LG models, this approach can now be employed to facilitate more advanced applications, such as the automated evolutionary design of engineering systems, and the modeling of larger more complex multi-domain dynamic systems, monitoring and design optimization of complex multi-domain mechatronic systems.

## III. Development of the Linear Graph Model for Mobile Robot

The physical four-wheel skid-skip mobile robot (Clearpath Husky) and the diagram of its subsystems are shown in Fig. 2. The LG model of the robot consists of various subsystems that encompass multiple physical domains and functions of the robotic system. The complete model includes the electrical subsystem, consisting of the DC motors powered by a voltage source (battery), the drivetrain subsystems, consisting of the front and rear axles and wheels for both of the independent left and right side powertrains, and the translational and rotational subsystems, representing the linear and rotational movements of the entire mobile robot.

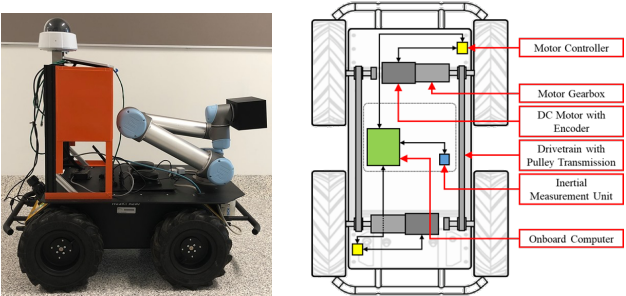

Fig. 3. Clearpath Husky mobile robotic system and the diagram of its subsystems.

### A. Electrical Subsystem

The Husky's electrical subsystem consists of two 24V brushed DC motors, each with a 78.71:1 gearbox reduction attachment, which transmit the output torque of the motors to the primary axles of the two independent drivetrains. Each DC motor drive module is modeled as a series LG circuit consisting of a controlled DC voltage source ($V_{S1}$), a D-Type resistive element ($R_1$), a T-Type inductive element ($L_1$), and a two-port transformer element representing the combined output of the motor torque constant and the gearbox reduction ($T_{ML}$).

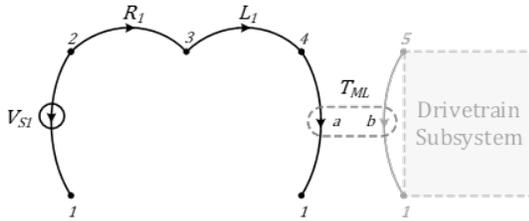

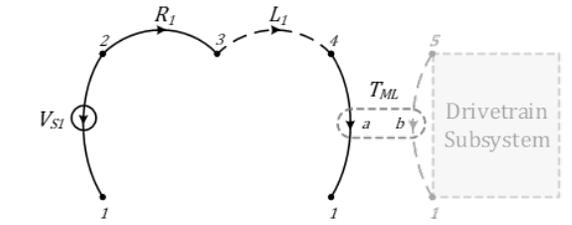

Fig. 4. LG Model and the Normal tree of the Husky robot electrical subsystem.

The equations produced by the LGtheory toolbox for the electrical subsystem based on the constitutive equations and the normal tree of the LG model are as follows:

The continuity equations for each passive branch are:

$$i_{R_1} = i_{T_{ML}} = i_{L_1} \quad (3)$$

The compatibility equation for each loop formed by the temporary inclusion of each passive link is:

$$V_{L_1} = V_{S1} - V_{T_{ML}} - V_{R_1} \quad (4)$$

### B. Drivetrain and Wheel Subsystem

The drivetrain subsystem of Husky consists of two independent drivetrains, each powered by its own DC motor. Each of these independent drivetrains consists of a primary axle, to which the DC motor is directly connected, and a secondary axle, which receives power from the primary axle via a 1:1 belt drive system. Each of these independent drivetrains power both of the wheels on their respective side of the vehicle. 4 is a diagram of the system configuration of Husky, with the drivetrain components and sensors used for data collection.

The present paper presents two methods for modeling the drivetrain subsystem. The first method, termed the expanded model, considers the dynamics of all the wheels on the vehicle separately and models the compliance and slippage of the belt drive system; and the second method, termed the simplified model, assumes the belt drive system is not flexible and models the inertias of the two axles of each drivetrain as a single element.

For the expanded model (here considering only the drivetrain on the left-hand side of the vehicle, as the other half of the drivetrain is modeled in the same manner), the second branch of the two-port element (coming from the corresponding motor of the electrical subsystem) splits at its upper node into four paths. The first two of these paths, which are set in parallel with the second port of the motor transformer, consist of elements that represent the parameters of the primary axle of the drivetrain. The first path contains an A-Type element ($J_{RL}$) and the second path consists of a D-Type element ($B_{RL}$) in series with two transformer elements ($TF_1$ and $TF_2$). The first path represents the combined inertial load of the wheel and shaft of the primary axle, whereas, the second path collectively represents all of the energy that is lost or transferred out of the drivetrain system. The D-Type element represents the energy dissipation from the system due to friction in the shaft bearings and also the friction of the wheels slipping on the driving surface, and the transformer elements represent the transmission of torque from the drivetrain subsystem to the translational and rotational dynamics subsystems of the Husky vehicle. The third and forth paths consist of a D-type ($B_{BeltL}$) element and a T-type ($K_{BeltL}$) element in parallel, representing the compliance and slip of the belt in the pulley transmission system. These paths then split at their shared second node into two more paths representing to the inertial load ($J_{FL}$) and the energy dissipation/transmission ($B_{FL}$, $TF_3$, and $TF_4$) of the secondary axle, similar to the primary axle.

Fig. 5 illustrates the aforementioned benefits of LG modeling over other modeling approaches. The structure of the LG model has a closer resemblance to the physical system than any other form of dynamic system modeling, such as a BG model. In particular, the LG model show how the elements representing the dynamics of each axle and overlay their respective wheels, while the parallel D-type and T-type elements closely resemble the belt that connects the two shafts.

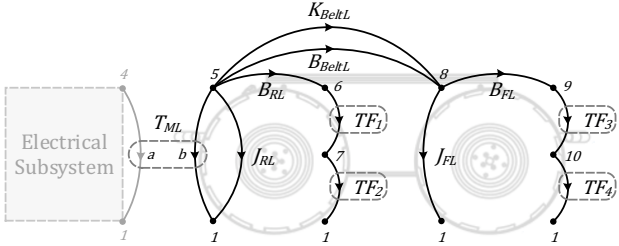

Fig. 5. LG model of the left-side drivetrain subsystem overlaid on the profile of the Husky robot.

For the simplified model, it is assumed that the belt is not flexible, thus eliminating the compliance between the primary and secondary axles of the drivetrain. With this change, the model can be further simplified by combining the inertia of the primary and secondary axels into a single element ($J_L$).

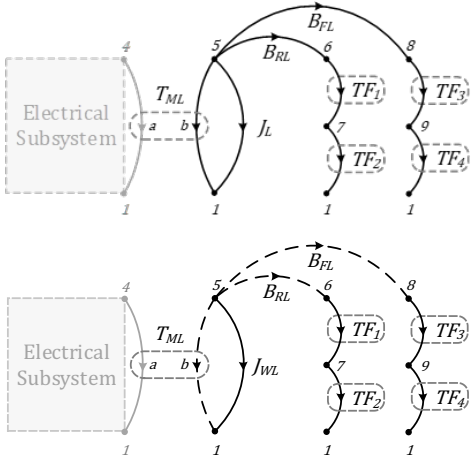

Fig. 6. Simplified LG model and the normal tree of the left-side drivetrain subsystem.

The equations produced by the LGtheory toolbox for the simplified drivetrain subsystem based on the constitutive equations and normal tree of the LG model are as follows:

The continuity equations for each passive branch are:

$$\tau_{J_{LW}} = -\tau_{B_{FL}} - \tau_{B_{RL}} - \tau_{T_{ML}}$$
$$\tau_{TF_1} = \tau_{TF_2} = \tau_{B_{RL}} \quad (5)$$
$$\tau_{TF_3} = \tau_{TF_4} = \tau_{B_{FL}}$$

The compatibility equations for each loop formed by the temporary inclusion of each passive link are:

$$\omega_{T_{ML}} = \omega_{J_{LW}}$$
$$\omega_{B_{RL}} = \omega_{J_L} - \omega_{TF_1} - \omega_{TF_2} \quad (6)$$
$$\omega_{B_{FL}} = \omega_{J_L} - \omega_{TF_3} - \omega_{TF_4}$$

*1) Drivetrain Transformer Equations*

The four transformer elements of the half drivetrain model shown in Fig. 7 represents the conversion of the wheel torque into the traction force that propels the vehicle linearly ($TF_{odd}$), and the conversion of the torque/traction force of the wheel into the moment that rotates the vehicle ($TF_{even}$). The equations that define these power conversions are determined next.

For $TF_{odd}$, the rotation of the wheels of the vehicle due to the power provided by the DC motors create a friction between the tread of the tire and the driving surface ($F_{F_i}$). This traction between the two surfaces creates a counter force on the vehicle ($F_{W_i}$), which propels the vehicle forward.

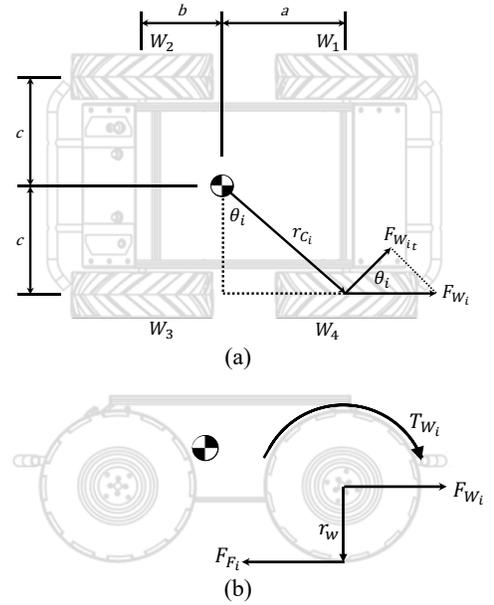

Fig. 7. Diagram of: (a) vehicle rotational moments, and (b) wheel tractive forces produced by the torque of the wheels.

Referring to the constitutive equations for a transformer, as given in Table II, $f_1$ is the generalized through-variable of the first port, representing the wheel torque, and $f_2$ is the generalized through-variable of the second port, representing the traction force of the wheels. In order to convert the wheel torque into a force, $f_1$ must be divided by the radius of the wheel ($r_W$), as:

$$TF_{odd} = \frac{1}{r_W} \quad (7)$$

Similarly, for $TF_{even}$, the traction force that propels the vehicle, generated by the torque of the DC motors, also produces the moment that rotates the vehicle.

Referring to the same constitutive equations for a transformer as before, $f_1$ is the generalized through-variable of the first port,

representing the wheel torque, and $f_2$ is the generalized through-variable of the second port, representing the moment applied on the vehicle. In this transformer, two conversions occur; first, the conversion of the wheel torque into traction force, then the conversion of the traction force into the rotational moment of the vehicle. Again, to create the traction force, the torque of the wheel must be divided by the radius of the wheel. To convert this traction force into rotational torque, the distances from the center of mass of the vehicle to the contact point of each wheel and the driving surface ($r_{c_i}$) must be found. The rotational torque of the vehicle produced by each wheel can then be determined by multiplying these distances by the tangential component of the force of each wheel (where (8) is "+" for right side wheels and "–" for left):

$$\text{TF}_{even} = \pm \cos(\theta_{W_i}) \cdot r_{C_i} \cdot \frac{1}{r_W} \quad (8)$$

Here $TF_2$ corresponds to the rear left wheel ($W_2$), $TF_4$ the front left ($W_1$), $TF_6$ the front right ($W_4$), and $TF_8$ the rear right ($W_3$), as shown in Fig. 7. Therefore, for $TF_2$ and $TF_8$ the following equations apply:

$$\theta_{W_2} = \theta_{W_3} = tan^{-1}\left(\frac{b}{c}\right) \quad (9)$$

$$r_{C_2} = r_{C_3} = \sqrt{(b)^2 + (c)^2} \quad (10)$$

Also, for $TF_4$ and $TF_6$ the following equations apply:

$$\theta_{W_1} = \theta_{W_4} = tan^{-1}\left(\frac{a}{c}\right) \quad (11)$$

$$r_{C_1} = r_{C_4} = \sqrt{(a)^2 + (c)^2} \quad (12)$$

### C. Mobile Robot Translational Dynamic Subsystem

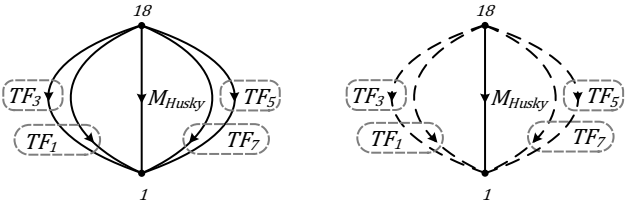

Fig. 8. LG model and the normal tree of the Husky robot translational dynamic subsystem.

The translational dynamics of the Husky robot are determined by the summation of the traction forces produced by the wheels of the vehicle. This is represented in LG form by placing the second ports of the odd numbered transformer elements in parallel with an A-Type element ($M_{Husky}$) representing the mass of the Husky vehicle. This configuration means that the forces produced by each tire will be summed in order to induce an acceleration in the vehicle when the wheels produce a non-zero resultant force, and allow the vehicle to remain stationary when the tire forces are balanced (i.e., the vehicle is stopped or turning on the spot).

The equations generated by the LGtheory toolbox for the translational dynamic subsystem of the mobile robot based on the constitutive equations and the normal tree of the LG model are as follows:

The continuity equation for each passive branch is:

$$F_{M_{Husky}} = -F_{TF_1} - F_{TF_3} - F_{TF_5} - F_{TF_7} \quad (13)$$

The compatibility equations for each loop formed by the temporary inclusion of each passive link are:

$$v_{TF_1} = v_{TF_3} = v_{TF_5} = v_{TF_7} = v_{M_{Husky}} \quad (14)$$

### D. Mobile Robot Rotational Dynamics Subsystem

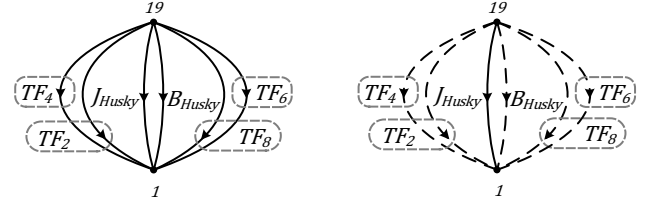

Fig. 9. LG model and the normal tree of the Husky robot rotational dynamic subsystem.

The rotational dynamics of the Husky robot are determined by the summation of the rotational moments produced by the wheels of the vehicle. This is represented in LG form by placing the second ports of the even numbered transformer elements in parallel with a D-Type element ($B_{Husky}$) and an A-Type element ($J_{Husky}$) representing the resistance to rotational movement and the inertia of the Husky vehicle, respectively. This configuration means that the torques produced by each tire will be summed in order to induce a rotation on the vehicle when the wheels produce a non-zero resultant moment, and allow the vehicle to remain stationary when the rotational moment is balanced (i.e., vehicle is not turning).

The equations produced by the LGtheory toolbox for the mobile robot rotational dynamics subsystem based on the constitutive equations and the normal tree of the LG model are given next.

The continuity equation for each passive branch is:

$$\tau_{J_{Husky}} = -\tau_{TF_2} - \tau_{TF_4} - \tau_{TF_6} - \tau_{TF_8} - \tau_{B_{Husky}} \quad (15)$$

The compatibility equations for each loop formed by the temporary inclusion of each passive link are:

$$\omega_{TF_2} = \omega_{TF_4} = \omega_{TF_6} = \omega_{TF_8} = \omega_{J_{Husky}}$$
$$\omega_{B_{Husky}} = \omega_{J_{Husky}} \quad (16)$$

### E. Complete Linear Graph Model of Mobile Robot

The complete LG model of the Husky robot is produced by combining the LG models of the subsystems presented in the previous section. This results in the LG model presented in Fig. 10, which encompasses the dynamics of the entire Husky system.

By evaluating the sets of equations produced by the various subsystem models, along with the corresponding equations for the right side powertrain, and the constitutive equations for each element, the LGtheory toolbox generates the following state-space model for the Husky system (*Note*: The "Husky" subscript has been replaced with "H"):

$$\frac{dx}{dt} = A \begin{bmatrix} \omega_{J_{LW}} \\ \omega_{J_{RW}} \\ v_{M_H} \\ \omega_{J_H} \\ i_{L_1} \\ i_{L_2} \end{bmatrix} + B \begin{bmatrix} V_{s1} \\ V_{s2} \end{bmatrix} \quad (17)$$

$$y = C \begin{bmatrix} \omega_{J_{LW}} \\ \omega_{J_{RW}} \\ v_{M_H} \\ \omega_{J_H} \\ i_{L_1} \\ i_{L_2} \end{bmatrix} + D \begin{bmatrix} V_{s1} \\ V_{s2} \end{bmatrix} \quad (18)$$

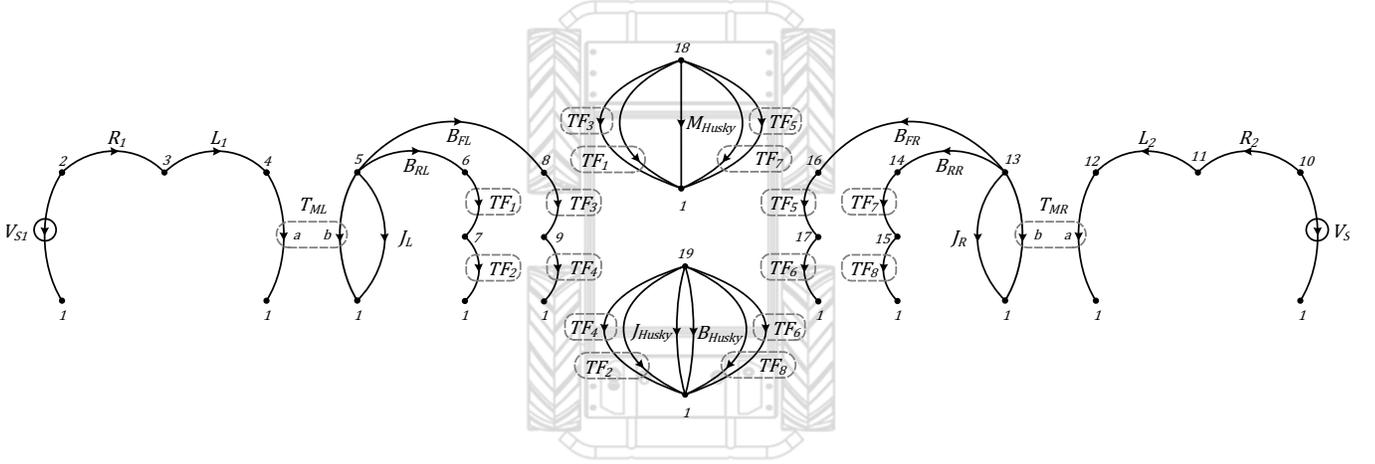

Fig. 10. Complete LG model of the Clearpath Husky Robot with the simplified drivetrain subsystems.

$$A = \begin{bmatrix} \frac{-B_{FL}-B_{RL}}{J_L} & 0 & \frac{B_{FL}TF_3+B_{RL}TF_1}{J_L} & \frac{B_{FL}TF_4+B_{RL}TF_2}{J_L} & \frac{T_{ML}}{J_L} & 0 \\ 0 & \frac{-B_{FR}-B_{RR}}{J_R} & \frac{B_{FR}TF_5+B_{RR}TF_7}{J_R} & \frac{B_{FR}TF_6+B_{RR}TF_8}{J_R} & 0 & \frac{T_{MR}}{J_R} \\ \frac{B_{FL}TF_3+B_{RL}TF_1}{M_H} & \frac{B_{FR}TF_5+B_{RR}TF_7}{M_H} & \frac{-B_{RL}TF_1^2-B_{FL}TF_3^2-B_{FR}TF_5^2-B_{RR}TF_7^2}{M_H} & \frac{-B_{FL}TF_3TF_4-B_{FR}TF_5TF_6-B_{RL}TF_1TF_2-B_{RR}TF_7TF_8}{M_H} & 0 & 0 \\ \frac{B_{FL}TF_4+B_{RL}TF_2}{J_H} & \frac{B_{FR}TF_6+B_{RR}TF_8}{J_H} & -B_{FL}TF_3TF_4-B_{FR}TF_5TF_6-B_{RL}TF_1TF_2-B_{RR}TF_7TF_8 & \frac{-B_{RL}TF_2^2-B_{FL}TF_4^2-B_{FR}TF_6^2-B_{RR}TF_8^2-B_H}{J_H} & 0 & 0 \\ -\frac{T_{ML}}{L_1} & 0 & 0 & 0 & -\frac{R_1}{L_1} & 0 \\ 0 & -\frac{T_{MR}}{L_2} & 0 & 0 & 0 & -\frac{R_2}{L_2} \end{bmatrix} \quad (19)$$

$$B = \begin{bmatrix} 0 & 0 \\ 0 & 0 \\ 0 & 0 \\ 0 & 0 \\ 1/L_1 & 0 \\ 0 & 1/L_2 \end{bmatrix} \quad (20)$$

$$C = \begin{bmatrix} 1 & 0 & 0 & 0 & 0 & 0 \\ 0 & 1 & 0 & 0 & 0 & 0 \\ 0 & 0 & 1 & 0 & 0 & 0 \\ 0 & 0 & 0 & 1 & 0 & 0 \end{bmatrix} \quad (21)$$

$$D = [0]_{4 \times 2} \quad (22)$$

The outputs of the state-space model are chosen to be the state-variables representing the rotational velocities of the left and the right wheels, and also the linear and rotational velocities of the mobile robot itself. The selection of these outputs allows for a comparison of these states with the corresponding Gazebo simulated and experimental state values of the robot, for the purposes of model validation and parameter estimation.

*F. Parameter Estimation using Genetic Algorithms*

GAs are a form of multi-point, population-based methods for simultaneously exploring multiple solutions to an optimization problem. Based on the same concepts as natural evolution, GAs reproduce and evolve members of a population, referred to as solutions, over many generations in order to obtain an optimized solution to a problem. Throughout this evolutionary process, new solutions inherit the beneficial characteristics from the successful solutions of past generations while also introducing new characteristics that may provide advantages over other solutions. Those solutions that are successful are more likely to reproduce, and thus, pass on their beneficial characteristics, while those that are less successful face the possibility of being purged from the population; this process tends to lead to a stronger population of solutions over many generations. GAs are stochastic in nature; therefore, they do not necessarily guarantee finding the most optimal solution to a problem, rather, this method is useful for more complex optimization problems that are difficult or infeasible to solve through mathematical means [16].

The present paper utilizes the GA capabilities of the Global Optimization Toolbox of MATLAB. The objective function of the parameter estimation GA is based on the sum of the absolute tracking errors in the $x$ and $y$ directions between the LG-based mobile robot simulation and the data obtained from Gazebo and from the physical experiments using the Husky robot.

$$obj = \sum |x_{data}(t) - x_{LG}(t)| + \sum |y_{data}(t) - y_{LG}(t)| \quad (23)$$

While most of the system parameters of the Husky robot can be determined from the manufacturer documentation or from the component datasheets, some system parameters that are based on specific environmental conditions must be calibrated for. These parameters, which will be calibrated using a GA, are the unknown damping constant values given in Table III, representing the losses of the drivetrain systems due to slip and friction ($B_{LW}, B_{RW}$), and the resistance to the rotational movement between the wheels of the Husky vehicle and the driving surface ($B_{Husky}$). Additionally, the GA will calibrate the coefficient value ($c$) of a simple multivariable function which is used to estimate the motor voltage signals from the recorded command velocity signals sent to the robot. Since real-time measurements of the motor voltages cannot be obtained at a sufficiently fast rate from ROS for the physical experiments, the following functions have to be utilized in order to estimate the voltage inputs to the state-space model:

$$V_{s1} = 24 \cdot (c \cdot v_t + 0.541 \cdot c \cdot v_r) \quad (24)$$

$$V_{s2} = 24 \cdot (c \cdot v_t - 0.541 \cdot c \cdot v_r) \quad (25)$$

Here, $v_t$ and $v_r$ are the command values for the translational and rotational velocities, respectively. It was found through experimentation with the command signals to the motor that a rotational velocity command which was numerically equivalent to a translational velocity command would result in 54.1% rotational speed of the Husky wheels; hence, the inclusion of the 0.541 values in equations (24) and (25).

Table III presents the known and unknown parameter values for the mobile robot state-space model.

TABLE I
STATE-SPACE PARAMETERS INCLUDING UNKNOWN VALUES

| Description | Parameter | Value | Units |
|---|---|---|---|
| Voltage Inputs | $V_{s1}, V_{s2}$ | ±24 | $V$ |
| Internal Motor Resistance | $R_1, R_2$ | 0.46 | $\Omega$ |
| Internal Motor Inductance | $L_1, L_2$ | 0.22 | $mH$ |
| Motor Torque Constant | $k_t$ | 0.044488 | $N \cdot m/A$ |
| Gear Ratio | $GR$ | 78.71 : 1 | Gear Ratio |
| Motor Transformer Ratio | $T_{ML}, T_{MR}$ | $k_t \times GR$ | $N \cdot m/A$ |
| Drivetrain Inertia | $J_{LW}, J_{RW}$ | 0.08 | $kg \cdot m^2$ |
| Drivetrain Damping | $B_{RL,FL,FR,RR}$ | Unknown | $rad/(N \cdot m \cdot s)$ |
| Power Conversion Transformer Ratios | $TF_{odd}$ | Equation (7) | |
| | $TF_{even}$ | Equation (8) | |
| Husky Mass | $M_{Husky}$ | 48.39 | $kg$ |
| Husky Rotational Damping | $B_{Husky}$ | Unknown | $rad/(N \cdot m \cdot s)$ |
| Husky Inertia | $J_{Husky}$ | 3.0556 | $kg \cdot m^2$ |

## IV. RESULTS AND DISCUSSION

### A. ROS Gazebo Simulation Environment

Gazebo is a physics-based 3D environment for simulating the rigid-body dynamics of robotic systems. Along with dynamic simulations, Gazebo can be used for simulating sensor readings, evaluating and training artificial intelligence (AI)-based control systems, and much more. Integration of Gazebo with ROS allows for exploiting the data transfer and communication capabilities which ROS facilitates. These capabilities facilitate recording of the command signals and sensor data of the robot using the following ROS topics: /tf, the transformation frames of the robot used for linear speed and trajectory; /imu_um7/data, the IMU data readings for rotational speed of the husky; /joint_states, the encoder readings of each wheel; and /husky_velocity_controller/cmd_vel, the command signals sent to the robot from the gamepad controller. ROS then provides this recorded data to be filtered for the command data and replayed as input to a Gazebo simulation in order to seek replicating the results.

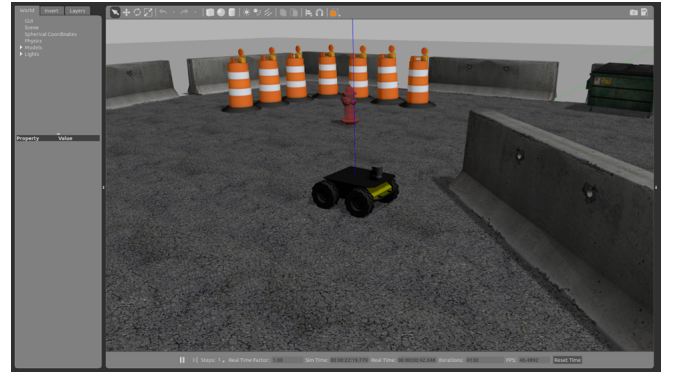

Fig. 11. Example of a Gazebo simulation environment.

### B. GA Estimation and Comparison of Results

Initial calibration of the Husky LG model was conducted using data collected from the physical robot while executing a circular maneuver. This maneuver involved an initial straight trajectory before entering a circular trajectory. Once a full revolution was complete, the Husky exited the circle again in a straight path.

Fig. 12 shows the performance of the GA-based parameter estimation procedure. The procedure used to calibrate the model was run with a population of 100 solutions, for a maximum of 100 generations, and with a crossover fraction of 0.5. Table IV gives the parameters that are estimated, the upper and lower bounds of their searches, and the results obtained which produced the optimal simulation results.

TABLE II
VARIABLES ADAPTED BY THE GA AND RESULTING VALUES

| Variable | $B_{FL}, B_{RL}, B_{RR}, B_{FR}$ | $B_{Husky}$ | $c$ |
|---|---|---|---|
| Upper Bounds | 1 | 1 | 0.75 |
| Lower Bounds | 100 | 100 | 1.00 |
| Results | 1.3016 | 12.8650 | 0.8961 |

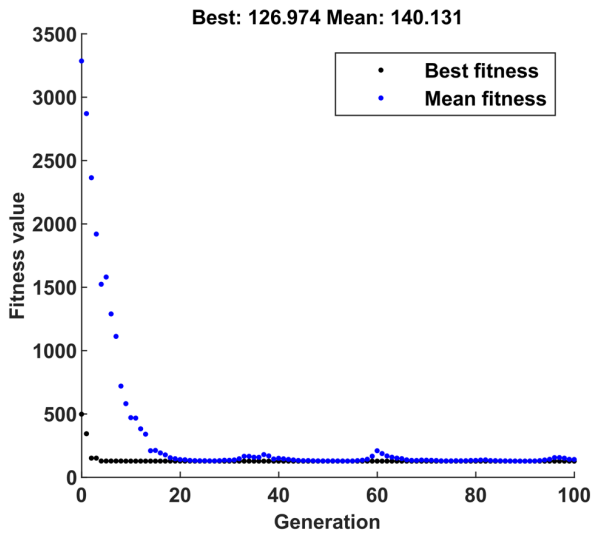

Fig. 12. GA evolution of parameter estimation.

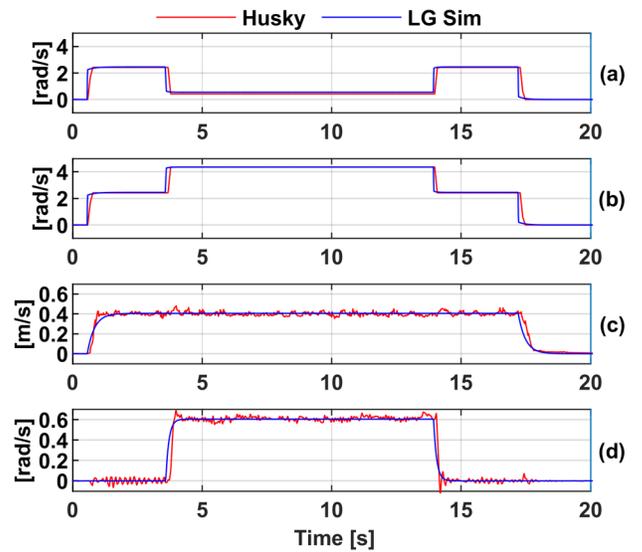

Fig. 13. Response of: (a) left and (b) right side wheel velocities; (c) linear and (d) rotational velocities of the Husky LG model and the measured Husky data.

Fig. 13 shows the response of the state-space outputs specified for the LG model against the corresponding measured data from the Husky vehicle for the duration of the circular maneuver. As can be seen, there is a strong conformance between the response of the simulated output states and the Husky sensor readings.

Fig. 14 shows the trajectory response and the tracking error of the calibrated LG model in comparison to the trajectory of the physical robot and the Gazebo simulated robot. As can be observed from this figure, the trajectory of the LG model closely agrees with the trajectory of the real robot, whereas, the trajectory of the Gazebo simulation differs significantly despite the same system inputs are used in the two simulation approaches. The bounds of the maximum tracking error of the LG-based simulation for this maneuver are:

$$|\bar{X}| \leq 0.1397 [m] \qquad |\bar{Y}| \leq 0.0819 [m]$$

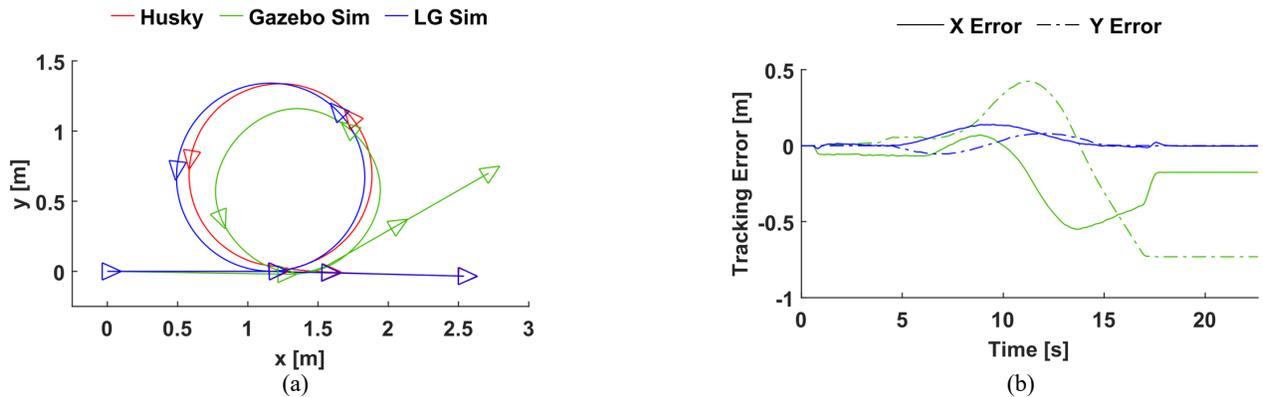

Fig. 14. (a) Trajectory, (b) Tracking error, of the LG and Gazebo simulations and experimental data for circle maneuver.

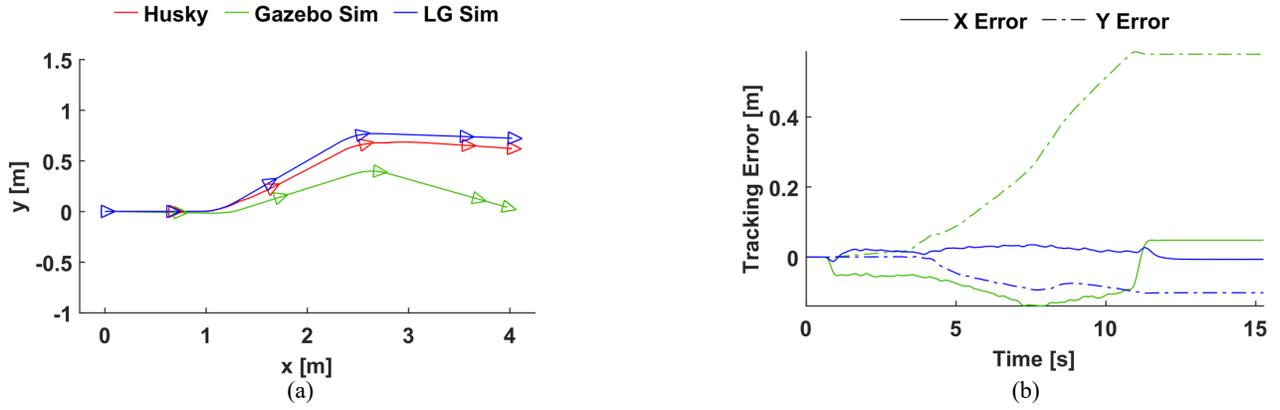

Fig. 15. (a) Trajectory, (b) Tracking error results of the LG and Gazebo simulations compared with experimental data for S-bend maneuver.

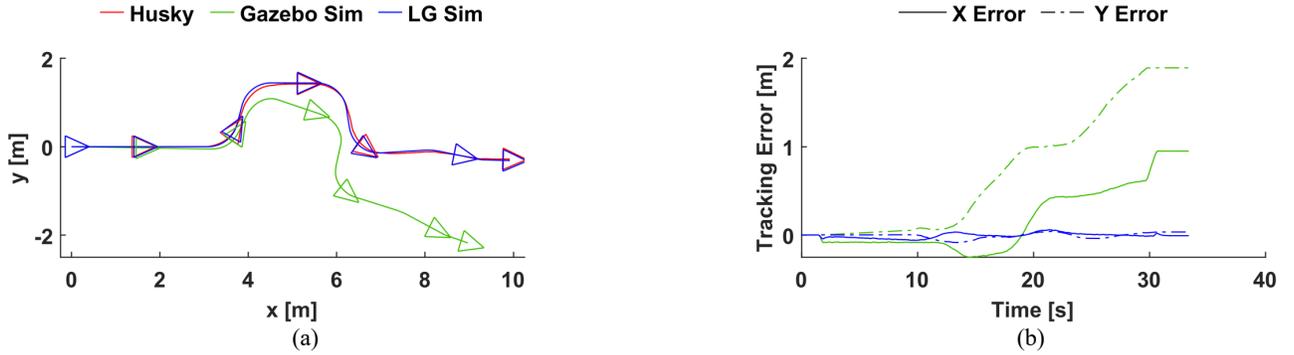

Fig. 16. (a) Trajectory, (b) Tracking error results of the LG and Gazebo simulations compared with experimental data for obstacle avoidance maneuver.

Fig. 15 shows the trajectory response and the tracking error of the pre-calibrated LG model and the Gazebo simulations in comparison with the response of the physical robot for an S-bend maneuver. As can be observed from this figure, the trajectory of the LG model closely agrees with the trajectory of the physical robot, while, again, the trajectory of the Gazebo simulation differs quite significantly. The bounds of the maximum tracking error of the LG-based simulation for this maneuver are:

$$|\bar{X}| \leq 0.0352 [m] \qquad |\bar{Y}| \leq 0.1027 [m]$$

Fig. 16 presents the trajectory response and the tracking error of the pre-calibrated LG model and the Gazebo simulations in comparison with the response of the physical robot for an obstacle avoidance maneuver. Again, the trajectory of the LG model closely follows the trajectory of the real robot; whereas, the trajectory of the Gazebo simulation differs significantly after the initial turn. The bounds of the maximum tracking error of the LG-based simulation for this maneuver are:

$$|\bar{X}| \leq 0.0613 [m] \qquad |\bar{Y}| \leq 0.0854 [m]$$

In each of the presented three maneuvers, the calibrated LG model provides a significantly more accurate trajectory response than the Gazebo Simulation. The primary cause of error for the Gazebo simulation is the excessive skidding, which can be observed during rotational movements of the Husky during simulation. For the circle maneuver, this excessive skidding resulted in a smaller, oval shaped trajectory, compared to the larger, circular trajectory of the LG model. Similarly, for the S-bend and obstacle avoidance maneuvers, the excessive skidding present in the gazebo simulation has resulted in skewing of the trajectory of the Gazebo robot. While some noticeable error is present in the LG model, it is much less than what is in the Gazebo simulations. The cause of this error is likely due to the complexity associated with modeling the dynamics of the interactions between the wheels of a skid-steer vehicle and the driving surface.

V. CONCLUSION

This paper presented a Linear Graph (LG)-based method of modeling the dynamics of a mobile robotic system, together with some background on the modeling approach of LG. The recently developed LGtheory MATLAB toolbox was utilized to automate the process of deriving the state-space model of a complex dynamic system such as the considered mobile robotic system. The genetic algorithm (GA) capabilities of the Global Optimization Toolbox were employed for estimating the unknown parameter values of the robot. The results of a comparison between a computer simulation of the state-space model for the mobile robot as generated by the LGtheory toolbox in comparison with a ROS/Gazebo-based simulation and experimental data collected while driving a physical Clearpath Husky mobile robot demonstrated and validated the accuracy of the developed modeling approach. The successful application of this modeling approach to a mobile robotic system, as demonstrated in the present work, provides further validation of both the proposed LG model and the custom software used to construct it.

ACKNOWLEDGMENT

This research was funded by the NSERC Discovery Program, grant number RGPIN-2017-05762.

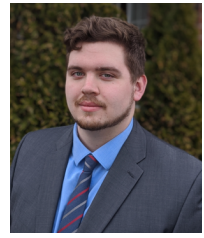
**Eric McCormick** received his B.Eng. (Honours) degree in mechanical engineering from Ontario Tech University, Oshawa, ON, Canada in 2018. He is currently pursuing his M.A.Sc. in mechanical engineering from Ontario Tech University. His research interests include dynamic system modeling using linear graph theory, automated design evolution, machine learning, and robotics.

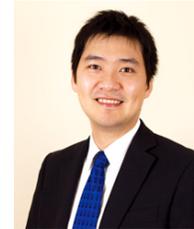
**Haoxiang Lang** received the Ph.D. degree from the Department of Mechanical Engineering, The University of British Columbia, Vancouver, BC, Canada, in 2012. Subsequently, he worked as a Postdoctoral Research Fellow in the Industrial Automation Laboratory of the University of British Columbia. Currently, he is an Assistant Professor in the Department of Automotive, Mechanical and Manufacturing Engineering, University of Ontario Institute of Technology. His research and development areas are Mechatronics, autonomous robotics, visual servoing, advanced controls, and machine learning.

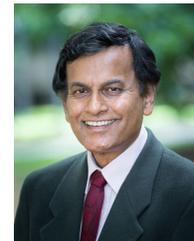
**Clarence W. de Silva** received the Ph.D. degrees from the Massachusetts Institute of Technology, Cambridge, MA, USA, in 1978, and the University of Cambridge, Cambridge, England, U.K., in 1998; the honorary D.Eng. degree from the University of Waterloo, Waterloo, ON, Canada, in 2008, and the higher doctorate of Sc.D. from the University of Cambridge in 2020. He has been a Professor of Mechanical Engineering and a Senior Canada Research Chair and NSERC-BC Packers Chair in Industrial Automation, University of British Columbia, Vancouver, Canada since 1988. He has authored 24 books and 555 technical papers, approximately half of which are in journals. Dr. de Silva is a Fellow of: ASME, IEEE, ASI, Canadian Academy of Engineering, and Royal Society of Canada.